\documentclass[10pt,conference]{IEEEtran}

\usepackage[utf8]{inputenc}
\usepackage{graphicx}
\usepackage{amsmath}
\usepackage{amssymb}
\usepackage{cite}
\usepackage{url}
\usepackage{multirow}
\usepackage{array}
\usepackage{booktabs}
\usepackage{algorithm}
\usepackage{algorithmic}
\usepackage{xcolor}
\usepackage{subcaption}
\usepackage{balance}
\usepackage{colortbl}

\DeclareMathOperator*{\argmin}{arg\,min}

\IEEEoverridecommandlockouts

\title{RAP: Retrieve, Adapt, and Prompt-Fit for Training-Free Few-Shot Medical Image Segmentation}
\author{*Zhihao Mao\textsuperscript{1}, *Bangpu Chen\textsuperscript{1}}

\begin{document}

\maketitle
\begingroup
\renewcommand\thefootnote{}
\footnotetext{\textsuperscript{1}School of Computer Science, China University of Geosciences, Wuhan, China. \textsuperscript{*}Equal contribution.}
\endgroup

\begin{abstract}
    Few-shot medical image segmentation (FSMIS) has achieved notable progress, yet most existing methods mainly rely on semantic correspondences from scarce annotations while under-utilizing a key property of medical imagery: anatomical targets exhibit repeatable high-frequency morphology (e.g., boundary geometry and spatial layout) across patients and acquisitions. We propose RAP, a training-free framework that Retrieves, Adapts, and Prompt-Fits Segment Anything Model 2 (SAM2) for FSMIS. First, RAP retrieves morphologically compatible supports from an archive using DINOv3 features to reduce brittleness to single-support choice. Second, it adapts the retrieved support mask to the query by fitting boundary-aware structural cues, yielding an anatomy-consistent pre-mask under domain shifts. Third, RAP converts the pre-mask into prompts by sampling positive points via Voronoi partitioning and negative points via sector-based sampling, and feeds them into SAM2 for final refinement—without any fine-tuning. Extensive experiments on multiple medical segmentation benchmarks show that RAP consistently surpasses prior FSMIS baselines and achieves state-of-the-art performance. Overall, RAP demonstrates that explicit structural fitting combined with retrieval-augmented prompting offers a simple and effective route to robust training-free few-shot medical segmentation.
\end{abstract}


\section{Introduction}
\label{sec:introduction}

Medical image segmentation is fundamental to clinical diagnosis, treatment planning, and disease monitoring~\cite{azad2024review}.
Accurate delineation of anatomical structures such as cardiac chambers, abdominal organs, and tumors enables quantitative analysis that directly impacts patient outcomes.
Deep learning methods have achieved remarkable success in this domain, yet they typically require large amounts of pixel-level annotations that are expensive and time-consuming to obtain.

Few-shot segmentation has emerged as a promising paradigm to address the annotation scarcity problem, where models learn to segment novel classes given only one or a few labeled support images~\cite{wang2019panet,ouyang2022self}.
However, existing few-shot methods face critical challenges in medical imaging.
First, they often require task-specific training on base classes, limiting their generalization to unseen domains and modalities~\cite{zhu2023few,lei2022cross}.
Second, the quality of segmentation heavily depends on the visual similarity between support and query images, which is problematic given the high variability in medical scans due to different scanners, protocols, and patient populations.
Third, most methods fail to exploit the strong anatomical priors inherent in medical images, where organs exhibit consistent shapes, positions, and topological structures across patients.
More broadly, retrieval-oriented systems suggest that external repositories can complement frozen models without retraining~\cite{li2025hierarchical,zheng2026heisd}.

We observe that medical images possess two distinctive properties that can be leveraged for training-free few-shot segmentation.
\textbf{(1) Boundary-centric characteristics}: The discriminative information in medical images concentrates around object boundaries, where intensity gradients and texture patterns differ significantly between foreground and background regions.
\textbf{(2) Strong anatomical priors}: Organs of the same category share similar shapes and spatial layouts, providing geometric constraints that can guide the segmentation process.

Based on these observations, we propose \textbf{RAP} (\textbf{R}etrieve, \textbf{A}dapt, and \textbf{P}rompt-Fit), a training-free framework for few-shot medical image segmentation.
RAP operates in three stages: (1) \textit{Retrieve} leverages DINOv3~\cite{siméoni2025dinov3} features to identify the most semantically similar support sample from a database; (2) \textit{Adapt} employs oriented chamfer matching guided by DINO-based semantic gating to align the support shape onto the query image, producing a preliminary mask; and (3) \textit{Prompt-Fit} generates geometrically-aware point prompts through Voronoi partitioning~\cite{aurenhammer1991voronoi}, which are fed into SAM2~\cite{ravi2024sam2} to obtain the final segmentation.
Unlike existing methods that require task-specific training or fine-tuning, RAP directly leverages pre-trained foundation models (DINOv3 and SAM2) without any gradient updates, requiring only a small database of annotated samples for retrieval.

Our main contributions are summarized as follows:
\begin{itemize}
    \item We propose RAP, a novel training-free framework for few-shot medical image segmentation that synergistically combines vision foundation models with geometric shape priors.
    \item We introduce an oriented chamfer matching algorithm with DINO-based semantic gating that enables robust shape adaptation from support to query images, effectively handling appearance variations across different domains.
    \item We design a Voronoi-based prompt generation strategy that produces geometrically-aware positive and negative points, significantly improving SAM2's segmentation quality in medical imaging scenarios.
    \item Extensive experiments on three medical imaging datasets demonstrate that RAP achieves competitive performance compared to state-of-the-art methods while requiring no training.
\end{itemize}

Upon acceptance, we will release our code and pre-built support databases to facilitate reproducibility and future research in training-free medical image segmentation.

\section{Related Work}
\label{sec:related_work}

\textbf{Few-shot Medical Image Segmentation.}
Few-shot segmentation aims to segment novel classes with limited labeled examples.
Prototype-based methods~\cite{wang2019panet,snell2017prototypical} extract class prototypes from support features and match them against query features.
SSL-ALPNet~\cite{ouyang2022self} introduces self-supervised learning to improve prototype quality, while RPT~\cite{zhu2023few} partitions foreground into multiple regions for comprehensive feature representation.
PATNet~\cite{lei2022cross}, IFA~\cite{nie2024cross}, and FAMNet~\cite{bo2025cross} address cross-domain challenges through domain-invariant features and frequency-aware matching.
However, these methods require task-specific training and struggle to generalize across different imaging modalities.

\textbf{Shape Priors in Medical Segmentation.}
Incorporating shape priors has been a long-standing approach to improve segmentation robustness~\cite{kervadec2019boundary}.
Recent methods integrate shape information through shape-aware loss functions~\cite{ma2021loss} or boundary attention mechanisms.
MAUP~\cite{ZhuYaz_MAUP_MICCAI2025} proposes a training-free approach using multi-center prompting with Voronoi-based region partitioning~\cite{aurenhammer1991voronoi}.
However, existing methods either require training to learn shape representations or lack explicit geometric alignment between support and query images.

\textbf{Foundation Models for Segmentation.}
SAM~\cite{kirillov2023segment} and SAM2~\cite{ravi2024sam2} have demonstrated impressive zero-shot segmentation capabilities by leveraging large-scale pre-training.
Several works have explored adapting SAM for medical imaging through prompt engineering~\cite{huang2025sam,wang2025osam}.
Self-supervised vision transformers like DINOv2~\cite{oquab2023dinov2} and DINOv3~\cite{siméoni2025dinov3} provide semantically rich features for retrieval and correspondence.
Our work combines DINOv3 for semantic guidance and SAM2 for segmentation, bridging them through geometric shape adaptation without any training.

\section{Preliminary}
\label{sec:preliminary}

\subsection{Problem Formulation}

We consider the one-shot medical image segmentation setting.
Given a support set $\mathcal{S} = \{(I_s, M_s)\}$ containing a single image $I_s \in \mathbb{R}^{H \times W}$ with its corresponding binary mask $M_s \in \{0, 1\}^{H \times W}$, and a query image $I_q \in \mathbb{R}^{H \times W}$, the goal is to predict the segmentation mask $\hat{M}_q$ for the query image.
In practice, we maintain a support database $\mathcal{D} = \{(I_s^{(i)}, M_s^{(i)})\}_{i=1}^{N}$ containing $N$ labeled samples, from which the most suitable support is retrieved for each query.

\subsection{Notations}

We denote query and support images as $I_q$ and $I_s$, with $M_s$ being the ground-truth mask of the support image and $\hat{M}_q$ being the predicted mask for the query.
The DINO feature maps are denoted as $\mathbf{F}_q, \mathbf{F}_s \in \mathbb{R}^{h \times w \times d}$, where $h \times w$ is the spatial resolution and $d$ is the feature dimension.
We use $E_q \in \mathbb{R}^{H \times W}$ to represent the edge map of the query image, $D_s \in \mathbb{R}^{H \times W}$ for the signed distance field computed from $M_s$, and $M_{\text{pre}} \in \{0, 1\}^{H \times W}$ for the preliminary mask obtained after shape adaptation.
The sets of positive and negative prompt points are denoted as $\mathcal{P}^+$ and $\mathcal{P}^-$, respectively.

\subsection{Foundation Models}

\textbf{DINOv3.}
DINOv3~\cite{siméoni2025dinov3} is a self-supervised vision transformer that extends DINOv2 with improved feature representations.
It produces semantically rich patch-level features that capture both local texture and global context.
Given an image $I$, DINOv3 extracts features $\mathbf{F} = \text{DINO}(I) \in \mathbb{R}^{h \times w \times d}$, where each spatial location encodes a $d$-dimensional descriptor.
These features exhibit strong correspondence properties, making them suitable for cross-image matching and retrieval tasks.

\textbf{Segment Anything Model 2 (SAM2).}
SAM2~\cite{ravi2024sam2} is an upgraded promptable segmentation model that extends SAM with improved architecture and training.
SAM2 takes an image and a set of prompts (points, boxes, or masks) as input and produces segmentation masks.
Given point prompts $\mathcal{P} = \{(p_i, l_i)\}$ where $p_i \in \mathbb{R}^2$ denotes the 2D coordinates and $l_i \in \{0, 1\}$ indicates positive or negative label, SAM2 outputs $\hat{M} = \text{SAM2}(I, \mathcal{P})$.
The quality of segmentation critically depends on the accuracy and placement of prompt points, which motivates our geometry-aware prompt generation strategy.

\section{Methodology}
\label{sec:methodology}

\subsection{Overview}

\begin{figure*}[t]
    \centering
    \includegraphics[width=\textwidth]{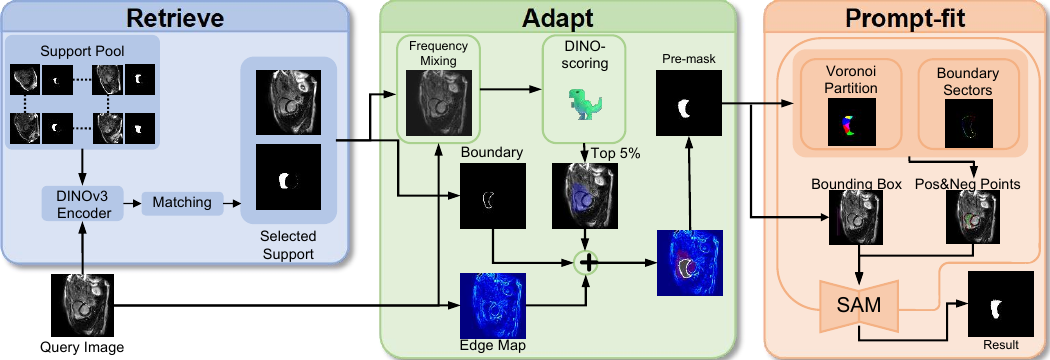}  
    \caption{Overview of the proposed RAP framework. Given a query image, we first retrieve the most similar support from a database using DINOv3 features (Retrieve). Then, we adapt the support shape to the query through oriented chamfer matching with semantic gating (Adapt). Finally, we generate Voronoi-based point prompts and feed them to SAM2 for segmentation (Prompt-Fit).}
    \label{fig:framework}
\end{figure*}

Fig.~\ref{fig:framework} illustrates the overall framework of RAP.
Given a query image $I_q$, our method operates in three stages.
\textbf{(1) Retrieve}: We extract global DINO features and retrieve the most semantically similar support sample $(I_s, M_s)$ from the database.
\textbf{(2) Adapt}: We employ oriented chamfer matching guided by DINO-based semantic gating to align the support mask shape onto the query image, yielding a preliminary mask $M_{\text{pre}}$.
\textbf{(3) Prompt-Fit}: We apply Voronoi partitioning on $M_{\text{pre}}$ to generate geometrically-distributed positive points, sample negative points from boundary sectors, and feed these prompts along with a bounding box to SAM2 for final segmentation.

\subsection{Retrieve Stage}
\label{sec:retrieve}

The retrieval stage aims to find the support sample whose anatomical structure best matches the query.
We leverage DINOv3 to extract semantically meaningful global descriptors.
For each image $I$, we compute the feature map $\mathbf{F} = \text{DINO}(I)$ and obtain the global descriptor by averaging over spatial locations:
\begin{equation}
    \mathbf{g} = \frac{1}{hw} \sum_{i,j} \mathbf{F}_{i,j}
    \label{eq:global_desc}
\end{equation}
For support images, we optionally compute a masked descriptor using only foreground regions to better capture target-specific features.

Given the query descriptor $\mathbf{g}_q$ and a database of support descriptors $\{\mathbf{g}_s^{(i)}\}_{i=1}^{N}$, we compute cosine similarities and retrieve the top-$k$ matches:
\begin{equation}
    \text{sim}(q, s^{(i)}) = \frac{\mathbf{g}_q \cdot \mathbf{g}_s^{(i)}}{\|\mathbf{g}_q\| \|\mathbf{g}_s^{(i)}\|}
    \label{eq:cosine_sim}
\end{equation}
In practice, we select the second-best match (rank-2) to avoid potential self-matching artifacts when query and database share similar sources.

\subsection{Adapt Stage}
\label{sec:adapt}

The adapt stage transfers the support mask shape to the query image while accounting for geometric variations.
This stage consists of three components: frequency-domain style adaptation, DINO-based semantic gating, and oriented chamfer matching.

\subsubsection{Frequency-domain Style Adaptation}

\begin{figure}[t]
    \centering
    \includegraphics[width=\columnwidth]{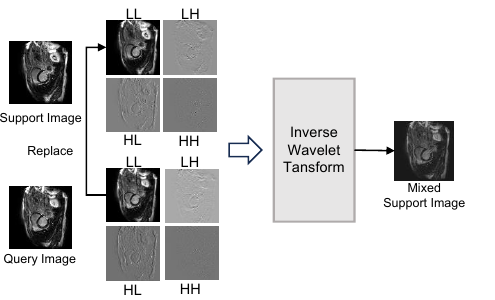}
    \caption{Frequency-domain style adaptation via wavelet transform. The support's low-frequency (LL) subband is replaced with the query's LL to align global appearance while preserving high-frequency edge details.}
    \label{fig:freq}
\end{figure}

To reduce domain shift between support and query images caused by different scanners or imaging protocols, we employ wavelet-based style transfer.
As illustrated in Fig.~\ref{fig:freq}, we decompose both images using discrete wavelet transform (DWT) into four subbands: LL (low-frequency approximation), LH, HL, and HH (high-frequency details).
The low-frequency subband captures global appearance such as brightness and contrast, while high-frequency subbands preserve edges and textures.
We replace the support's LL subband with the query's LL subband and apply inverse wavelet transform to obtain a style-adapted support image:
\begin{equation}
    \tilde{I}_s = \text{IDWT}(\text{LL}_q, \text{LH}_s, \text{HL}_s, \text{HH}_s)
    \label{eq:freq_adapt}
\end{equation}
This operation transfers the query's global appearance to the support while preserving the support's edge structures, which are critical for shape matching.

\subsubsection{DINO-based Semantic Gating}

To narrow down the search space for shape matching, we compute a semantic similarity map between support and query.
We first cluster the support mask into $K$ regions using K-Means on DINO features:
\begin{equation}
    \{R_k\}_{k=1}^{K} = \text{KMeans}(\mathbf{F}_s[M_s > 0], K)
    \label{eq:clustering}
\end{equation}
For each region, we compute a prototype by averaging features within the region and measure its similarity to query features:
\begin{equation}
    S_k(i,j) = \frac{\mathbf{F}_q(i,j) \cdot \mathbf{c}_k}{\|\mathbf{F}_q(i,j)\| \|\mathbf{c}_k\|}
    \label{eq:region_sim}
\end{equation}
where $\mathbf{c}_k$ is the prototype of region $R_k$.

We select the top-$K'$ regions with highest average similarity and aggregate their maps to form a gating mask $G$:
\begin{equation}
    G = \bigcup_{k \in \text{top-}K'} \mathbb{1}[S_k > \tau_s]
    \label{eq:gating}
\end{equation}
where $\tau_s$ is a threshold determined by the $q$-th quantile of similarity values.
This mask identifies candidate regions in the query where the target structure is likely to appear.

\subsubsection{Oriented Chamfer Matching}

We compute the query edge map $E_q$ by fusing multi-scale Laplacian of Gaussian (LoG) responses with Sobel gradients:
\begin{equation}
    E_q = w_{\text{LoG}} \cdot \max_{\sigma} |\text{LoG}_\sigma(I_q)| + w_{\text{grad}} \cdot \|\nabla I_q\|
    \label{eq:edge_map}
\end{equation}
where $\sigma \in \{1, 2, 4, 8\}$ are the LoG scales.

From the support mask $M_s$, we extract boundary points $\{(x_i, y_i)\}$ along with their normal directions $\{\theta_i\}$ by computing tangent vectors along the contour.
These points form the shape template to be matched against the query.

To enable orientation-aware matching, we partition the query edge pixels into $K_\theta$ angular bins based on their gradient directions.
For each bin $k$, we compute a distance transform $\text{DT}_k$ on the corresponding edge pixels:
\begin{equation}
    \text{DT}_k(x, y) = \min_{(x', y') \in E_q^{(k)}} \|(x, y) - (x', y')\|
    \label{eq:directional_dt}
\end{equation}
where $E_q^{(k)}$ denotes edge pixels in angular bin $k$.

We search for the optimal transformation $(t_x, t_y, s, r)$ (translation, scale, rotation) that minimizes the oriented chamfer distance:
\begin{equation}
    \mathcal{C}(t_x, t_y, s, r) = \frac{1}{|\mathcal{B}|} \sum_{(x_i, y_i, \theta_i) \in \mathcal{B}} \text{DT}_{k(\theta_i + r)}(T(x_i, y_i))
    \label{eq:chamfer}
\end{equation}
where $\mathcal{B}$ is the support boundary template, $T(\cdot)$ applies the transformation, and $k(\theta)$ maps angle to bin index.
The search is constrained within the gating mask $G$ to improve efficiency and robustness.

The optimal transformation is applied to the support mask to obtain the preliminary mask:
\begin{equation}
    M_{\text{pre}} = T^*(M_s) \cap G
    \label{eq:premask}
\end{equation}

\subsection{Prompt-Fit Stage}
\label{sec:prompt}

The prompt-fit stage converts the preliminary mask into effective SAM2 prompts through geometry-aware sampling.

\subsubsection{Voronoi-based Positive Points}

Inspired by recent works on region-based prompting~\cite{zhu2023few,ZhuYaz_MAUP_MICCAI2025}, we apply Farthest Point Sampling (FPS) on $M_{\text{pre}}$ to obtain $N_v$ seed points that are maximally spread:
\begin{equation}
    \mathcal{V} = \text{FPS}(M_{\text{pre}}, N_v)
    \label{eq:fps}
\end{equation}
These seeds induce a Voronoi partition of the mask region, where each cell $V_i$ contains pixels closest to seed $v_i$.

For each Voronoi cell, we select its centroid as a positive prompt point:
\begin{equation}
    \mathcal{P}^+ = \left\{ \text{centroid}(V_i) \mid i = 1, \ldots, N_v \right\}
    \label{eq:pos_points}
\end{equation}
This strategy ensures that positive points are geometrically distributed across the entire target region, providing comprehensive coverage for SAM2.

\subsubsection{Sector-based Negative Points}

We partition the boundary of $M_{\text{pre}}$ into $N_s$ angular sectors centered at the mask centroid.
Within each sector, we sample negative points from the exterior region based on DINO dissimilarity scores:
\begin{equation}
    \mathcal{P}^- = \bigcup_{j=1}^{N_s} \argmin_{p \in \text{Sector}_j \setminus M_{\text{pre}}} S_{\text{DINO}}(p)
    \label{eq:neg_points}
\end{equation}
where $S_{\text{DINO}}(p)$ is the semantic similarity at point $p$.
This ensures negative points are distributed around the object boundary with low similarity to the foreground.

\subsubsection{SAM2 Inference}

We combine the positive points $\mathcal{P}^+$, negative points $\mathcal{P}^-$, and a bounding box $\mathcal{B}_{\text{box}}$ derived from $M_{\text{pre}}$ as the complete prompt set:
\begin{equation}
    \hat{M}_q = \text{SAM2}(I_q, \mathcal{P}^+, \mathcal{P}^-, \mathcal{B}_{\text{box}})
    \label{eq:final_sam}
\end{equation}
The bounding box provides a spatial prior that constrains SAM2's attention to the region of interest, while the point prompts offer fine-grained foreground/background guidance.

\section{Experiments}
\label{sec:experiments}

\subsection{Datasets}

We evaluate our method on three widely-used medical image segmentation benchmarks spanning different modalities and anatomical structures.
\textbf{Abd-MRI}~\cite{kavur2021chaos} consists of 20 cases of abdominal MRI scans from the ISBI 2019 Combined Healthy Abdominal Organ Segmentation (CHAOS) challenge, with annotations for four abdominal organs: liver, left kidney (LK), right kidney (RK), and spleen.
\textbf{Abd-CT}~\cite{ABD-CT} contains 20 cases of abdominal CT scans from the MICCAI 2015 Multi-Atlas Labeling Beyond the Cranial Vault (BTCV) challenge, with the same four organs annotated.
\textbf{Card-MRI}~\cite{zhuang2018multivariate} comprises 45 cases of cardiac MRI scans from the MICCAI 2019 Multi-Sequence Cardiac MRI Segmentation Challenge, with three cardiac structures annotated: left ventricle blood pool (LV-BP), left ventricle myocardium (LV-MYO), and right ventricle (RV).

\subsection{Evaluation Metrics}

Following prior works~\cite{ZhuYaz_MAUP_MICCAI2025,zhu2023few}, we adopt the Dice Similarity Coefficient (DSC) as the primary evaluation metric: $\text{DSC} = \frac{2|P \cap G|}{|P| + |G|} \times 100\%$, where $P$ and $G$ denote the predicted and ground-truth masks, respectively.
All experiments are conducted under the 1-way 1-shot setting to simulate the scarcity of labeled data in medical scenarios.

\subsection{Implementation Details}

We employ the pre-trained DINOv3-ViT-L/16~\cite{siméoni2025dinov3} as the feature extractor and SAM2-ViT-H~\cite{ravi2024sam2} as the segmentation backbone.
All images are resized to $512 \times 512$ for DINO feature extraction.
For oriented chamfer matching, we use $K_\theta = 8$ angular bins, scales $s \in \{0.6, 0.7, \ldots, 1.4\}$, and rotations $r \in \{-20^\circ, -10^\circ, 0^\circ, 10^\circ, 20^\circ\}$.
The DINO gating threshold is set to the 90th percentile of similarity values.
For Voronoi-based prompt generation, we sample $N_v = 6$ positive points via FPS and $N_s = 8$ negative points from boundary sectors.
During retrieval, we select the rank-2 match instead of rank-1 to avoid potential self-matching when query and support slices originate from the same patient volume in leave-one-out evaluation.
All experiments are conducted on a single NVIDIA RTX 3090 GPU with 24GB memory using PyTorch.

\subsection{Compared Methods}

We compare RAP with several state-of-the-art few-shot medical image segmentation methods.
The training-based methods include PANet~\cite{wang2019panet}, SSL-ALPNet~\cite{ouyang2022self}, RPT~\cite{zhu2023few}, PATNet~\cite{lei2022cross}, IFA~\cite{nie2024cross}, and FAMNet~\cite{bo2025cross}, all of which require training on source medical domains.
For training-free comparison, we include MAUP~\cite{ZhuYaz_MAUP_MICCAI2025}, which leverages DINOv2 and SAM without gradient updates.

\section{Results and Discussion}
\label{sec:results}

\subsection{Comparison with State-of-the-art}

Table~\ref{tab:abd_results} presents the quantitative comparison on Abd-MRI and Abd-CT datasets.
Our RAP achieves the best average Dice scores of 69.16\% on Abd-MRI and 69.56\% on Abd-CT, outperforming all compared methods.
Notably, RAP surpasses the recent MAUP~\cite{ZhuYaz_MAUP_MICCAI2025} by 2.07\% and 2.10\% on the two datasets, respectively, validating the effectiveness of our shape-guided approach.

\begin{table}[t]
    \centering
    \caption{Quantitative comparison (Dice \%) on Abd-MRI and Abd-CT datasets.}
    \label{tab:abd_results}
    \resizebox{\columnwidth}{!}{
    \scriptsize
    \begin{tabular}{l|cccc|c|cccc|c}
        \toprule
        \toprule
        \multirow{2}{*}{\textbf{Method}} & \multicolumn{5}{c|}{\textbf{Abd-MRI}} & \multicolumn{5}{c}{\textbf{Abd-CT}} \\
        \cmidrule{2-11}
        & Liver & LK & RK & Spleen & Mean & Liver & LK & RK & Spleen & Mean \\
        \midrule
        \rowcolor{gray!15} PANet & 39.24 & 26.47 & 37.35 & 26.79 & 32.46 & 40.29 & 30.61 & 26.66 & 30.21 & 31.94 \\
        \rowcolor{pink!15} SSL-ALPNet & 70.74 & 55.49 & 67.43 & 58.39 & 63.01 & 71.38 & 34.48 & 32.32 & 51.67 & 47.46 \\
        \rowcolor{orange!15} RPT & 49.22 & 42.45 & 47.14 & 48.84 & 46.91 & 65.87 & 40.07 & 35.97 & 51.22 & 48.28 \\
        \rowcolor{yellow!15} PATNet & 57.01 & 50.23 & 53.01 & 51.63 & 52.97 & 75.94 & 46.62 & 42.68 & 63.94 & 57.29 \\
        \rowcolor{lime!15} IFA & 50.22 & 35.99 & 34.00 & 42.21 & 40.61 & 46.62 & 25.13 & 26.56 & 24.85 & 30.79 \\
        \rowcolor{violet!15} FAMNet & 73.01 & 57.28 & \textbf{74.68} & 58.21 & 65.79 & 73.57 & 57.79 & 61.89 & 65.78 & 64.75 \\
        \rowcolor{cyan!15} MAUP & 78.16 & 58.23 & 72.34 & 59.65 & 67.09 & 78.25 & 59.41 & \textbf{71.80} & 60.38 & 67.46 \\
        \rowcolor{green!20} \textbf{RAP (Ours)} & \textbf{79.84} & \textbf{60.28} & 72.06 & \textbf{64.46} & \textbf{69.16} & \textbf{80.16} & \textbf{60.62} & 69.54 & \textbf{67.93} & \textbf{69.56} \\
        \bottomrule
        \bottomrule
    \end{tabular}
    }
\end{table}

Table~\ref{tab:card_results} shows the results on the Card-MRI dataset.
RAP achieves an average Dice of 75.27\%, improving upon MAUP by 2.14\%.
The improvement is particularly significant on the challenging LV-MYO structure (+1.27\%) and RV (+3.58\%), where our oriented chamfer matching effectively captures the thin myocardium boundaries.

\begin{table}[t]
    \centering
    \caption{Quantitative comparison (Dice \%) on Card-MRI dataset.}
    \label{tab:card_results}
    \resizebox{\columnwidth}{!}{
    \begin{tabular}{l|ccc|c}
        \toprule
        \toprule
        \textbf{Method} & \textbf{LV-BP} & \textbf{LV-MYO} & \textbf{RV} & \textbf{Mean} \\
        \midrule
        \rowcolor{gray!15} PANet & 51.42 & 25.75 & 25.75 & 36.66 \\
        \rowcolor{pink!15} SSL-ALPNet & 83.47 & 22.73 & 66.21 & 57.47 \\
        \rowcolor{orange!15} RPT & 60.84 & 42.28 & 57.30 & 53.47 \\
        \rowcolor{yellow!15} PATNet & 65.35 & 50.63 & 68.34 & 61.44 \\
        \rowcolor{lime!15} IFA & 50.43 & 31.32 & 30.74 & 37.50 \\
        \rowcolor{violet!15} FAMNet & 86.64 & 51.82 & 76.26 & 71.58 \\
        \rowcolor{cyan!15} MAUP & 88.36 & 52.74 & 78.29 & 73.13 \\
        \rowcolor{green!20} \textbf{RAP (Ours)} & \textbf{89.92} & \textbf{54.01} & \textbf{81.87} & \textbf{75.27} \\
        \bottomrule
        \bottomrule
    \end{tabular}
    }
\end{table}

\subsection{Ablation Study}

We conduct ablation studies on the RV structure of Card-MRI to validate the contribution of each component in RAP.
As shown in Table~\ref{tab:ablation}, each module progressively improves the segmentation performance.

\begin{table}[t]
    \centering
    \caption{Ablation study on Card-MRI (RV structure). OCM: Oriented Chamfer Matching, SG: Semantic Gating, VP: Voronoi-based Prompts.}
    \label{tab:ablation}
    \resizebox{\columnwidth}{!}{
    \begin{tabular}{ccc|c}
        \toprule
        \toprule
        \textbf{OCM} & \textbf{SG} & \textbf{VP} & \textbf{Mean Dice (\%)} \\
        \midrule
        \rowcolor{gray!15} & & & 78.29 \\
        \rowcolor{pink!15} \checkmark & & & 80.17 \\
        \rowcolor{orange!15} \checkmark & \checkmark & & 81.02 \\
        \rowcolor{green!20} \checkmark & \checkmark & \checkmark & \textbf{81.87} \\
        \bottomrule
        \bottomrule
    \end{tabular}
    }
\end{table}

\textbf{Effect of OCM.}
Adding Oriented Chamfer Matching improves Dice from 78.29\% to 80.17\% (+1.88\%), demonstrating the importance of orientation-aware shape alignment.

\textbf{Effect of SG.}
DINO-based Semantic Gating further improves to 81.02\% (+0.85\%) by constraining the search space and preventing false matches.

\textbf{Effect of VP.}
Voronoi-based Prompts achieve 81.87\% (+0.85\%) through comprehensive geometric coverage of the target region.

\subsection{Qualitative Analysis}

\begin{figure}[t]
    \centering
    \includegraphics[width=\columnwidth]{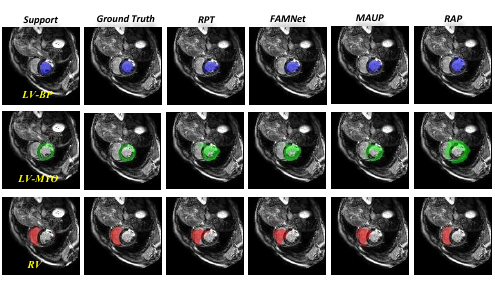}
    \caption{Qualitative results on Card-MRI for three cardiac structures. Our method produces more accurate boundaries, especially for the thin myocardium (LV-MYO).}
    \label{fig:qualitative}
\end{figure}

Fig.~\ref{fig:qualitative} presents qualitative comparisons on the Card-MRI dataset for three cardiac structures (LV-BP, LV-MYO, and RV).
Our RAP produces more accurate boundaries compared to baseline methods, especially for the thin myocardium (LV-MYO) where precise boundary delineation is critical.
The Voronoi-based prompts effectively guide SAM2 to segment complete structures without missing thin regions or including spurious areas.

\subsection{Discussion}

We clarify that ``training-free'' refers to requiring no gradient-based optimization or weight updates during inference.
While RAP utilizes a pre-built support database for retrieval, this database only stores raw images and masks without learned representations, distinguishing our approach from methods that require task-specific training on source domains.

RAP offers several key advantages.
First, it enables immediate deployment to new domains without retraining.
Second, the shape-guided approach is robust to appearance variations across different scanners and protocols.
Third, the oriented chamfer matching effectively handles gradient direction information for subtle boundaries.

Regarding the support database, our experiments use leave-one-out evaluation where each test case retrieves from remaining samples.
The retrieval mechanism is robust to database variations: since DINOv3 features capture semantic similarity rather than exact pixel matching, morphologically compatible supports can be retrieved even from small databases (20-45 cases in our benchmarks).

However, performance may degrade for rare anatomical structures or pathological cases that deviate significantly from normal anatomy in the database.

\section{Conclusion}
\label{sec:conclusion}

We presented RAP, a training-free framework for few-shot medical image segmentation that synergistically combines vision foundation models with geometric shape priors.
Our method introduces oriented chamfer matching with DINO-based semantic gating for robust shape adaptation, and Voronoi-based prompt generation for geometry-aware SAM2 guidance.
Experiments on three datasets demonstrate that RAP achieves competitive performance compared to training-based methods while requiring no task-specific optimization.
Future work includes extending RAP to 3D volumetric segmentation by leveraging inter-slice consistency.

\section*{Acknowledgments}

\bibliography{ref/bibliography}
\bibliographystyle{IEEEtran}

\end{document}